# Notion of Explainable Artificial Intelligence - An Empirical Investigation from A User's Perspective

AKM Bahalul Haque, LUT University , Finland, bahalul.haque@lut.fi.

A.K.M. Najmul Islam, LUT University , Finland, najmul.islam@lut.fi.

Patrick Mikalef, Norwegian University of Science and Technology, Norway, patrick.mikalef@ntnu.no.

**Abstract**

The growing attention to artificial intelligence-based applications has led to research interest in explainability issues. This emerging research attention on explainable AI (XAI) advocates the need to investigate end user-centric explainable AI. Thus, this study aims to investigate user-centric explainable AI and considered recommendation systems as the study context. We conducted focus group interviews to collect qualitative data on the recommendation system. We asked participants about the end users' comprehension of a recommended item, its probable explanation, and their opinion of making a recommendation explainable. Our findings reveal that end users want a non-technical and tailor-made explanation with on-demand supplementary information. Moreover, we also observed users requiring an explanation about personal data usage, detailed user feedback, and authentic and reliable explanations. Finally, we propose a synthesized framework that aims at involving the end user in the development process for requirements collection and validation.

## 1   Introduction

Recommendation systems are becoming an increasingly prevalent phenomenon in our daily lives. These recommendation systems use various machine learning models to suggest suitable content to the end users (Samek et al., 2019; Laato et al., 2022). Besides recommendation systems, AI-based applications can also be found in our smartphones as predictive text input, corporate organizations, national defense agencies, and many other places. (Samek et al., 2019;

Ehsan et al., 2021). However, much metadata is generated due to the increased usage of recommendation systems in product purchases, movies, music streaming, trip recommendations, etc. The system again uses this vast metadata to provide and rectify the recommendations. In this way, the recommendation systems are now able to personalize their suggestions for individual users. However, as these types of systems are becoming more and more prevalent, user concerns about the trust, transparency, and fairness of these systems are also increasing (Haque et al., 2023; Laato et al., 2022).

Recommendation systems use complex machine learning models which are considered black box since it is difficult to understand their working procedure (Samek et al., 2019; Adadi & Berrada, 2018). The models also do not reveal any information that lay people can understand. Therefore, the end users find it hard to trust a system that is not transparent and understandable to them (Stitini et al., 2022; He et al., 2016). This situation resulted in a growing demand for explainability in recommendation systems (AI-based content suggestions) (Gerlings et al., 2020; Gunning et al., 2019). It is to be mentioned that, in this study, we will use a recommendation system and AI-based content suggestions interchangeably. Gerlings et al., (2020) identified that if users understand how an AI works in content suggestion, it will also help shape their perception of AI (Gerlings et al., 2020). Therefore, explainable AI (XAI) can be vital in shaping user perception of trust and transparency in AI-based systems. In this paper, we have used several other terms along with explainability such as interpretability, and comprehensibility (Adadi & Berrada, 2018). Though they differ slightly in their literal meaning, scientists and researchers often use them interchangeably (Adadi & Berrada, 2018). Interpretability of a system refers to human understandability of a system where the user can clearly see the working procedure and study the mathematics behind it (how the inputs and outputs are related) (Adadi & Berrada, 2018; Gilpin et al., 2018). Similarly, comprehensibility also refers to the understandability and interoperability of the AI-based system (Piltaver et al., 2014).

XAI is defined as "Given an audience, an explainable Artificial Intelligence produces details or reasons to make its functioning clear or easy to understand" (Arrieta et al., 2020). Moreover, Arrieta et al. (2020) have discussed that explanations of various XAI systems depend entirely on the target audience (Arrieta et al., 2020). Therefore, explanations targeted for various actors related to the system will vary. For example, explanations for system developers and designers are given to make them understand the inherent working principle of the system to assist them with debugging, troubleshooting, or improving system performance. On the other hand, the explanations can help end users comprehend in layman's terms how specific content is suggested (Arrieta et al., 2020; Rebera & Lapedriza, 2019). Earlier research on XAI from the recommendation system perspective has focused on the social media recommendation trust, fake news detection, explanation related to music recommendation, hate speech recommendation, etc. In addition, most of the earlier studies on recommendation systems from the XAI perspective are either primarily technical or do not largely target lay users ( Abdul et al., 2018; Dhanorkar et al., 2021; Guidotti et al., 2018). Since recommendation systems are being used mainly by lay users, such as in e-commerce, social media, media, and entertainment, etc., the need for explainability of black box nature of AI systemts is also increasing. If this black box nature is reduced and the system is made explainable, the users can better understand the system, have more trust, and increase transparency (Haque et el., 2023; Dhanorkar et al., 2021). Hence, there is a growing concern among scholars to explore XAI from a non-technical lay users' perspective (Rebera & Lapedriza, 2019; Faraj et al., 2018). Such explorations can help the AI scholars and developer community to build a more user-centric, trustworthy, transparent, and understandable system for the end users (Haque et al., 2023). Exploring the topic from the end users' perspective can also reveal the need to include end users in the XAI development process (Waardenburg & Huysman, 2022; Ribera & Lapedriza, 2019). Therefore, the abovementioned reasons advocate the need to explore XAI from the end users' point of view. Consequently, we define our research objective as: "understanding the nature of explanations that end users require from an XAI system."

To address this research objective, in this study, we have collected empirical data using focus group interviews. We have adopted a qualitative approach to investigate the end users' perception of XAI from the recommendation system context (Smithson, 2000; Krueger & Casey, 2002). We have conducted focus group interviews for our paper, including 30 participants divided into 5 focus groups. Moreover, we have presented three scenarios to each focus group: an E-commerce product recommendation system, a housing recommendation, and a movie recommendation system (demo presented in Figure 1). Each group discussion lasted 60 to 90 minutes using the scenarios presented in Figure 1. The interview data were transcribed and analyzed by all the authors of this study (Guidotti et al., 2018). During the analysis phase, we used inductive coding and thematic analysis (Fereday & Muir-Cochrane, 2006). The analysis has three primary themes which are (i) User defined Explanation, (ii) Explanation regarding users' personal data, and (iii) Peer-reviewed explanation. The findings suggest three implications which are: (i) considering end users' requirements in XAI development, (ii) Explanation Verifiability, and Authenticity, and (iii) Theoretical Framework for End-user Inclusion and Requirement Validation for XAI Development

The rest of the paper is organized as, in the next section (section 2), we have discussed the background work related to XAI in general, human-centric XAI, and recommendation systems. Next in section 3, we have detailed a discussion on research methods and data collection. We have reported the findings in detail in section 4. Later we analyzed the findings and discussed the implications in section 5. Section 6 contains the limitations, followed by the paper's conclusion in section 7.

## 2       Explainable Artificial Intelligence

XAI systems aim to interpret the algorithms and explain the decision-making procedure to the users. Several explanation approaches and strategies have been presented to make AI systems more explicable, particularly for ML algorithms (Lamy, 2019; Vilone & Longo, 2021). The more complex the model, the more difficult it is to interpret. Often, there is a tradeoff between interpretability and accuracy (Bell et al., 2022). We can divide explainable techniques into two groups based on the complexity of ML algorithms: intrinsically explainable and post hoc explainable (Molnar, 2020). Algorithms such as decision trees are considered to be inherently interpretable but have less accuracy. However, algorithms that require post-hoc explanations are more complex and have higher accuracy (Guidotti et al., 2018). These algorithms are also termed the Blackbox model because, without proper explanation, it is tough to understand their working procedure. The post-doc explanation uses a separate set of techniques to perform a reverse engineering process to provide the needed explanations without altering or even knowing the inner workings of the original model (Guidotti et al., 2018). In addition, researchers use various visualization techniques and tools to make complex models (i.e., deep neural networks) understandable. However, these visualization techniques are also not a complete package to explain AI-based systems to the end users (Arrieta et al., 2020). Therefore, investigation on non-technical end user-centric XAI is needed.

The concept of interpretability specifies that understanding an AI model unfolds in two ways. One is to understand how a model behaves comprehensively, and another is to know how a prediction is made (Samek & Müller, 2019). Both concepts and most explainable AI techniques available today primarily focus on the technical aspects. Therefore, we can see that technical aspects of explainability methods have been largely investigated to date. However, since AI users are mostly lay users nowadays and the number is increasing rapidly, it is vital to focus on the non-technical explanation of AI. The end users' demand for explanation is also increasing with time (Gerlings et el., 2020). Therefore, an investigation of XAI purely from a non-technical end-user perspective can potentially assist in developing a proper user-centric XAI system (Jin et al., 2019). In this case, the recommendation system due to its wide usability among lay users can be a suitable example that can be studied and investigated more extensively from user centric XAI perspective (Samek et al., 2019; Haque et al., 2023).

## 2.1 Human-Centric XAI from Recommendation System Perspective

Miller (2019) articulated a comprehensive description of XAI from a social science perspective, especially regarding human comprehension and knowledge creation when people interact with any system (Miller, 2019). An algorithmic tradeoff exists in the case of explainable AI systems. Yu et al. (2020) investigated the algorithmic tradeoffs between fairness and accuracy from a system designer's perspective (Yu et al., 2020). Yu et al., (2020) also discussed that intelligent systems are becoming more human-centered as time passes and the need for algorithmic fairness is also increasing from the end user's perspective. However, among the explainable user-centric systems, some consider user reviews and behavioral analysis (Hernandez-Bocanegra & Ziegler, 2021; He et al., 2015). These systems try to convey an explanation that includes user reviews and product purchase patterns

Furthermore, along with user reviews, if users' personal preference and item attribute is added as an explanation, it can increase explainability of the system (Cheng et al., 2029). In this case, the visual detail of the item helps with users' prediction accuracy after the explanation is received. Samih et al. (2019) proposed an approach to improve the recommendation system using the acknowledged graph method; however, the authors did not consider the user-centric approach (Samih et al., 2019). Ooge et al. (2022) investigated user trust in recommendations provided by e-learning systems (Ooge et al., 2022). The authors analyzed the need for explanations in recommendation systems and later studied how explanations alongside recommendations impact user trust in the system.

Various explainable recommendation systems (Content-based, collaborative, and demographic filtering) induce trust at different levels. The explanation also varies according to the recommendation system being used (Liao et al., 2022). The explainable recommendation system has been explored using reinforcement learning and collaborative filtering (Wang et al., 2018), reinforcement learning and policy-guided approach (Xian et al., 2019), embedding-based methods, and easy-to-interpret attention network (Wang et al., 2018), heterogeneous knowledge base embeddings (Ai et el., 2018), etc. Therefore, XAI in a recommendation syste has been explored mostly from a technical perspective such as using various machine learning algorithms and understanding the trust of various types of filtering algorithms used in recommendation systems. Considering this fact, exploring explainable recommendation systems from non-technical end users' perspective is a timely topic and requires investigation. Therefore, in this study we have used three recommendation system scenarios to collect user opinions through focus group interview session. The end user opinion about explainable recommendation systems will help to design and develop human-centric XAI in the future (Zhang & Chen, 2020; Haque et al., 2023; Miller, 2019). Research Method

## 3      Study Design and Context

In this study, we conducted a focus group interview to understand users' perceptions of the recommendation system (Smithson, 2000; Krueger & Casey, 2002). We showed the participants three recommendation systems scenarios (Figure 1) and asked them about their comprehensibility. We wanted to know the extent to which users need the explainability and what are their requirements for making a system explainable. While discussing the scenarios, we referred them to popular websites like Amazon, Netflix, and Airbnb. The scenarios were designed by keeping these websites in mind. Therefore, the references helped the participants to understand the scenario better. Moreover, we informed the users that the opaqueness of our scenario is similar to the reference websites we mentioned here. We wanted to know from the participants what type of information they expect from the system along with a content recommendation. In addition, we also wanted to know how the explanation could be convincingly provided to them. As mentioned earlier, we have shown three different scenarios. The purpose of showing three scenarios is not to put them in a confusing situation but to make the participants understand what a recommendation system can look like in different domains. The scenarios also helped them to connect with the reference websites we mentioned earlier. Before conducting the sessions, we performed a pilot study to test the approach's feasibility and possible improvements. The peer reviewers were asked to review and comment on the relevance, interview scenario, approach, questions, and stimulus. The whole strategy was again modified based on their feedback. For example, in the pilot study we found that if the users are presented with some reference websites, they can easily relate to our topic of discussion. Some websites tell the users about recommending similar products based on previous purchase decisions, recommending movies based on previously watched categories, etc. Moreover, we found that users need some altealternativeds for explainability such as interpretability, hints, explanation, etc. Therefore, in our focus group discussion, we included these findings. In addition, we have asked domain experts and researchers to validate the questionnaire and approach we followed.

### 3.1   Participants and Focus Group Sessions

#### 3.1.1   Study participants

Regarding the number of focus group interviews, the minimum and maximum participant numbers are under debate in the literature, however, it is reasonable to have anything from 4 to 12 people in the focused group interviews (Çokluk et al., 2011). Such experiments typically require a relatively small sample size. In addition, participants beyond ten may disrupt group dynamics, and conversations may become out of hand (Goss, & Leinbach, 1996; Çokluk et al., 2011). Thus, we recruited 30 participants and distributed them among five focus groups. The participants were selected from different educational backgrounds and professions to ensure a wide variety of data collection. The participants were selected randomly. We invited the participants verbally and also via an open email and social media invitation.

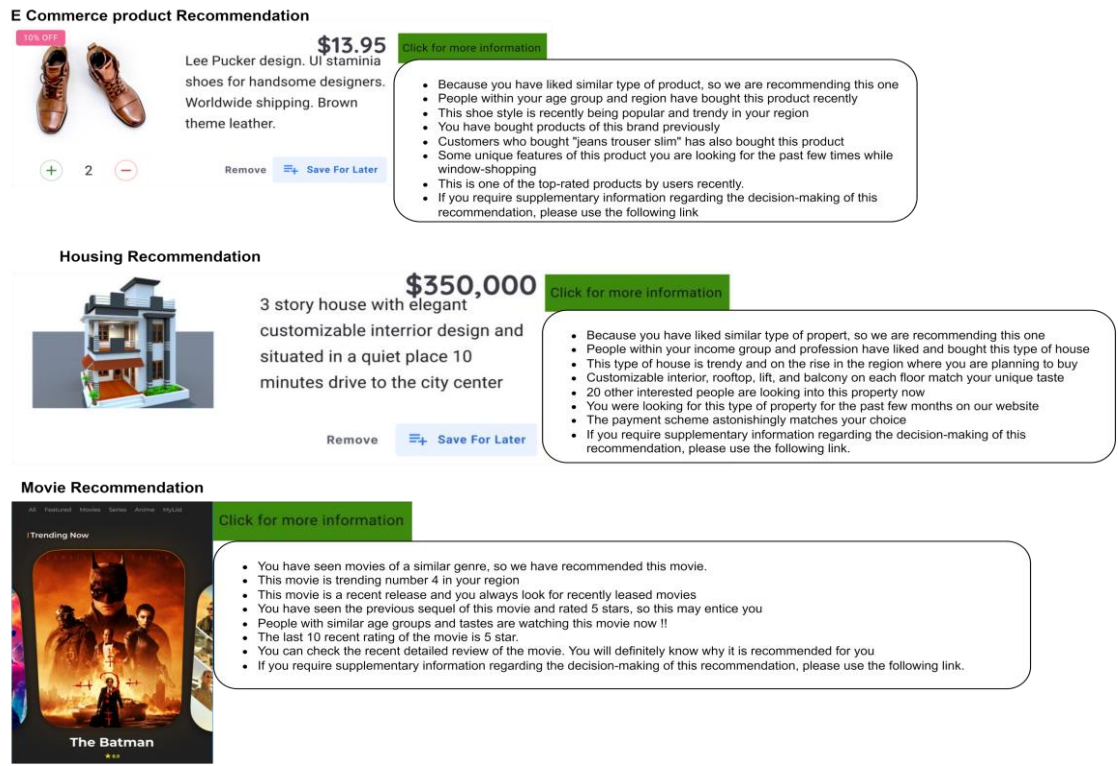

Figure 1.  The Explainable Recommendation System Scenarios

### 3.1.2  Focus group sessions

The focus group session started by welcoming the participants, and the facilitator explained why we were conducting this session. After this segment, the participants shared their opinion on regularly used recommendation systems. Later, the facilitator presented the recommendation scenarios as highlighted in figure 1, and asked the participants about their perceptions of the items recommended in these scenarios. The participants started their discussion and shared their opinion among themselves. Then, the participants were asked about their view on why that specific item was recommended and which attributes they think are vital here. The facilitator asked, " Why is that specific hotel recommended to you." Most importantly, the facilitator brought up the topic of explainability and whether they have any specific requirements. Here a facilitator used these questions "Which information do you like to see with the recommendations? What type of information? Think of some examples/ideas on how the information can be delivered/shown to you. If you think the information provided by the service provider is not enough, what would you do ?" After this question, the participants were given time to discuss it among themselves. In between these question, we also used some alternative words as found in the pilot study. The researchers were taking notes of discussions during the whole session. These notes were helpful during the coding and thematic analysis. Finally, the facilitator asked the last question about the "nature" or "way" of presenting the explanations to the users. At the end of the session, the participants were presented with a summary of the whole session and asked to comment on or add anything if they wanted.

## 3.2 Data Collection and Analysis

The interview took place via web-based audio call, and each session lasted approximately 1.5 hours. Before conducting the interview, we collected explicit consent from the participants regarding the purpose of the interview and how the data will be processed. We did not use any personal data while analyzing the interview data. The interview sessions were recorded and transcribed for later analysis. The assistants also took notes, so essential points raised during the interview were highlighted for later use. The data analysis followed a two-step process. In the first step, we analyzed the recorded transcripts and identified the codes. In the next step, we used inductive coding and thematic analysis (Fereday & Muir-Cochrane, 2006); Braun & Clarke, 2006). According to Braun & Clarke, (2006), thematic analysis is about identifying similar patterns so that they could be grouped together and made sense. These similar patterns are consolidated to answer the research question. Therefore, in our case, the interview data was open-coded to label the initial codes. The labels are presented as the 1st order codes shown in Figure 2. It helped the researchers to find initial themes. Later, similar labels are grouped together and added under a specific category (2nd order codes shown in Figure 2). The similar categories were then grouped together into aggregated themes (shown in Figure 2). Consequently, we identified three primary themes which are mentioned in the thematic analysis section.

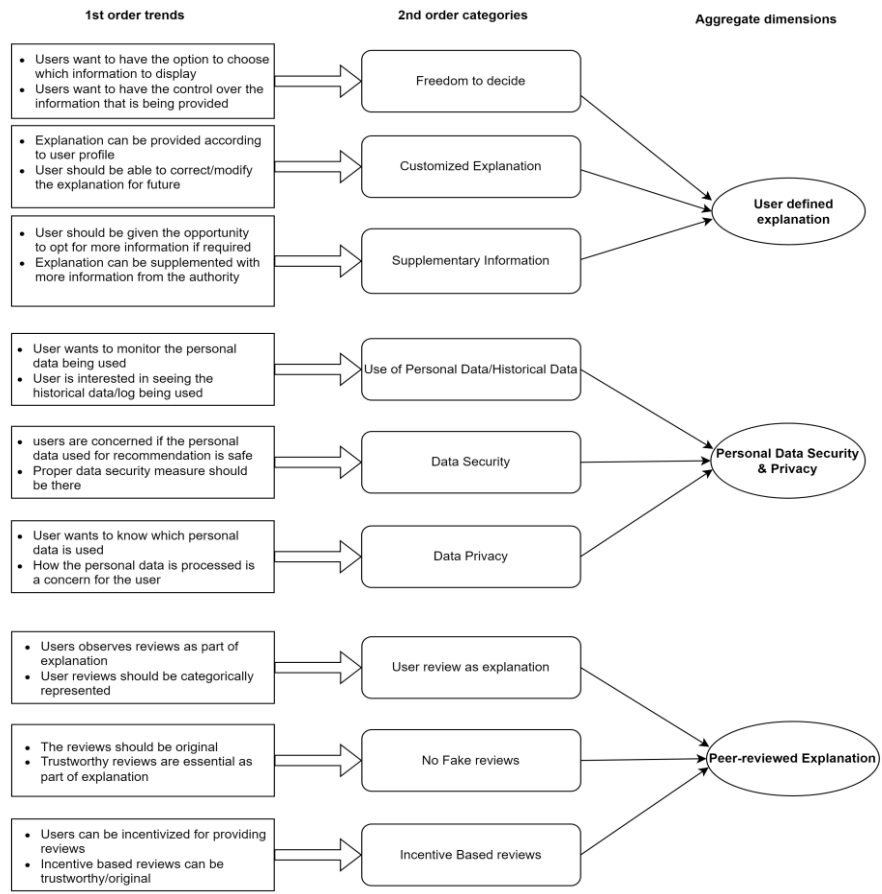

Figure 2.      Thematic analysis of the Focus Group Interview Data

## 3.3      Validity and Reliability

The group discussion data were examined using the inductive technique by applying a thematic analysis (Braun & Clarke, 2006). Thematic analysis is a procedure that involves conducting an in-depth and organized examination of qualitative data. Afterward, thse information was compiled and triangulated according to the different data types and attributes (Golafshani, 2003). In this study, we have used investigator triangulation. The primary author was responsible for supervising the entire data analysis process, beginning with the formulation of test questions. After that, the remaining investigators went back through the data analysis process and double-checked it. This process was iteratively done to ensure the integrity of the data. Moreover, after the discussion, the notes and crucial points were discussed with the participants to ensure that anything is missing or not. In this way, we validated and ensured the reliability of the collected data. In addition, before the interview, several preliminary questionnaires were developed to assess the reliability of the questionnaires. The participants of the study provided their responses voluntarily, and their data anonymity was ensured throughout the research.

## 4 Results and Findings

This section outlines the findings after analyzing the focus group interview data. The findings show a thematic analysis that addresses the research objectives of this study. Moreover, primary themes are divided into sub-themes for better understanding. The focus group interview demographics are summarized in Table 1. The demographic information highlights the participants in this study are mostly male (n=18) and remaining female (n=12). The study participants mainly include students and others are job holders and entrepreneurs.

| Characteristics | Specifics |
|---|---|
| Participants | 30 |
| Total Focus Groups | 5 |
| Facilitators | 1 |
| Validators | Researchers (3), Domain Experts (2) |
| Age Group | 19-35 years |
| Gender | Male = 18, Female = 12 |
| Employment Status | Job holder = 9; Student = 20 (Undergraduate = 10, Graduate = 15) Entrepreneur =1 |
| Interview Duration | 90 minutes (Average time) |

Table 1.     Focus group interview demographics

### 4.1     Thematic analysis

Thematic analysis of the interview data highlights that the nature of the explanation that end-user needs revolves around three primary themes which are: Tailor made explanations, Personal data security and privacy-related explanations, and Peer-reviewed explanations. These are explained in the following sections.

### 4.1.1 Tailor-Made Explanations

**Freedom to decide**

The participants discussed different recommendation systems that they come across daily. After presenting the scenarios to the participants, the facilitator asked them what they thought about the recommended items and how they wanted them to be explained. The participants discussed their need for explanation, especially when buying or viewing products online. Sometimes, they receive a promotional email if they register on any website or buy a product. However, the participants were interested in receiving an explanation with those emails. In most cases, the participants mentioned that they want to have the opportunity to give preferences before receiving any explanation of promotional or product recommendation emails. Otherwise, they may receive explanations they need help understanding, or it will be too much for them. This observation shows that users need more freedom and control over the information being provided to them. For example, one of the participants shared,

"I should be able to decide how much information is provided to me. Too much information is not good all while I am using the system. I should be provided with the opportunity to choose what I wish to see".

In this discussion, the issue of information overload has surfaced. Most participants said that providing users with too much information can lead to information overload, hindering the system's comprehensibility. Therefore, the user's should have control over the amount of information provided as explanation is needed. One of the participants said,

"As I have said previously, options should be there to get more information from the system. I do not want to be overloaded with a lot of unnecessary information". ….. "Indeed, we do not want to be overwhelmed with tons of information as an explanation; however, it is equally important that we all the necessary information about any product we want to buy."

Along with the controlled information delivery to the user, it is crucial to keep track of the explanation that is sufficient for the user. Though this task is tedious since the amount of required information varies from user to user. It can be considered an essential design requirement while developing a system. This finding also suggests that including end users in the development stage of AI-based systems can help to identify the required amount of explanation.

**Customized Explanation**

According to the participants' discussion, the recommendation systems should facilitate the user to tailor the explanations according to their needs. Everyone wants the system to act according to their demand. The focus group discussion revealed that users primarily wish to choose whether the system should display the explanation. They think it is highly unlikely that users will always require and expect an explanation. Though accommodating all the features can be very complex and challenging to attain, the users want the facility to customize the explanations to some extent. In this way, the whole system will be more user-centric.

For example, the users want to have individual user profile-based explanations. In any case, if there are multiple system users, one user should not receive any explanation based on another user's data. As one of the participants said,

> "If multiple users are using my computer, a mechanism should be there to track which recommendation/product is accessed/searched by whom. The reason is simple; I do not want to have any recommendation just because one of my siblings was window shopping from my laptop".

In addition to the user control over choosing the explanation format, users want to have the facility to inform the system if it is showing the wrong explanation. The algorithmic explanations are generally logical; in most cases, it needs to be more comprehensible for lay users. Therefore, users discuss that explanations should be less about algorithmic technicality and more about human comprehension. In that case, there should be an option in the interface where the user can provide their feedback and make a correction or modification. This feature will enable the recommender system to learn from this feedback in the future. The users believe this feature will give them more control over the system. One of the participants said,

> "Right now, I do not have any control of the information that I see on the website. As a user of any system, I think I should be able to customize or tweak some of the things. I am not talking about the whole system, but at least the bits and pieces".

From our observation, apart from the recommender system algorithm, the interface significantly impacts the perception of an explainable system. Therefore, the recommender system interface should be equipped with user requirements, for example, explanation choice, modification, feedback, etc. Later, the system can consider the user-defined corrections and provide better reasoning next time. It is reflected in the following interview quote,

> "If I find the recommendation correct, I should be able to correct the system. For example, I am searching for a house to buy, but none of the recommendations I got are suitable for me. In this case, the system should take my feedback so that I get to see exactly what I want next time."

**Supplementary Information**

The discussion revealed that apart from the explanation provided by the system, users were also interested in requesting more information. The participants believed that if users could get more info instantly from the system upon request, it would increase their comprehension of the system. One of the participants said,

"While using the system, I might require more information for clarification. The system should be responsible enough to let me do that. Therefore, I should be able to contact the concerned authority or system administrator, or any responsible entity. The option should be there for convenience."

Users think there may be a lack of information in the explanation. In addition, depending on the platform they are using, requesting more information is a "should have" feature. For example, one of the participants said,

> "For hotel/AirBnB recommendations, in my opinion, comparative information should be there. Comparison between different accommodations can present a clear picture".

Supplementary information will help the user to understand the options that are provided. However, the supplementary information can be related to the product, recommendation procedure, or both. In the latter case, the explanation presented to the users should be strictly from lay users' perspective because, from our study we have seen, it is very unlikely that the users will be experts in understanding algorithms.

**Explanation Accessibility, Availability, and Presentation Format**

Participants' responses highlighted accessibility and availability of explanation. If any information is not readily available, it should be notified to the user. Moreover, users want to have easy access to the explanations. One of the participants said,

> "The website should provide easy access to the information I require. If the system has explanation features, they should be available promptly, and I should be able to surf the required information easily. There should be no hurdle such as completing any forms, mandatory registration process, or paying any charges."

In addition, the system should also ensure the availability of the required explanation. If the explanation is provided to the user in a consistent format, it can increase the understandability of the recommendations being provided to the user. To emphasize this fact, the participant explicitly mentioned,

> "The explanation should be consistent. Meaning that the explanation in the same website/system should be presented in a similar format. Finding different visualization formats on the same website will confuse me, and It will be difficult for me to grasp them."

Easily accessible and available information as an explanation enhances the comprehensibility of the system. It also increases the reliability of the system. Our observation shows if the explanations are easily accessible and available in a specific format, the users tend to find them more understandable. Therefore, if the XAI system is developed according to the user requirements with a proper explanation format, it can impact the system's reliability.

4.1.2    Data security and privacy related Explanation

**Use of Personal Data**

Focus group discussion reveals that when participants receive a suggestion in a system, they are willing to know about their data processing procedures. In addition to the procedures, the participants want to know which personal data is being used to process the recommendations. The participants also highlighted that from various sources, they know these recommendation systems use personal data for recommendation purposes. Still, they are not clear about it since the system does not show anything regarding that as an explanation. Therefore, these user requirements generally complement the demand for explanations in XAI. One of the participants said,

> "I have heard from various news portals, blogs, and videos that AI is using our data. To be honest, it is scary and concerning. So, if they use our data, I think we have the right to know which data they collect and whether they are well protected. If such information is available to us along with explanations, we can be reassured about it."

Users think that all websites collect their data when they visit them because they come across the option of collecting user consent. In addition to brief information regarding an explanation, the participants discussed that they must know which data is being used for a specific item recommendation. For example, according to one of the participants,

> "If my information is used, the explanations should include which information is collected and why they are collected. I would be thrilled to see how my information is used to recommend me the product of my choice."

Furthermore, users think if their usage log is shown as an explanation, in some cases, it will make more sense. The participants believe that a historical data usage log can add new insights along with explanations provided by the recommended system. One participant said,

> "Our browsing history can be provided to us in a summarized format in the form of graphs or charts or any other understandable pictures. Reading a lot of text may be boring, but pictures can describe it. If any system is transparent in this way, I will definitely trust the system."

The findings from the group discussion reveal that explanations can help increase a system's transparency and induce user trust.

**Data Security and Privacy**

The participants deliberately started discussing personal data security and privacy issues in the wings. While the users addressed personal data usage used in the recommendation system, their concern shifted to data security privacy. Participants have learned about security breaches from social media and other news platforms. These sources have incited the thought about their data security and whether appropriate data security measures have been taken. One of the participants started the discussion with this opinion:

> "If the authority is continuously processing my data and my usage log, it is important for me to know. I believe as a user, we should be aware whether proper security measures have been taken".

Therefore, as part of the explanation, users intend to have the security measures adopted by the authority. They believe it will help them decide whether to trust the system.

In addition to security measures, data privacy is also a significant concern. Participants' discussion shifted toward data privacy along with security issues. Recent data privacy issues are a concern for users. To protect data privacy and consent management, the participants said that the explanation should include information regarding the consent management of personal data usage and data privacy and regulations (i.e., GDPR).

### 4.1.3 Peer-reviewed and Trustworthy Explanation

**User Review as Explanation**

This issue surfaced during the user discussion while they were talking about checking the reviews of any content. The participants strongly highlighted that users' reviews are vital to the explanation. Most participants agreed that they consider users' reviews while making the decision. One participant said,

> "Whenever I search for anything online, I tend to see the user reviews before going for it. Generally, products with more reviews pop up at first. For me, user reviews can also be considered part of the explanation."

Additionally, they outlined that users' reviews should not be merely how many stars (i.e. five stars, four stars, etc.) any content has received. Some of the participants suggested that user reviews should be detailed. The discussion highlighted an instrumental observation: user reviews/feedback help accept or reject the recommended product. One of the participants said,

> " Whenever I am searching for a product online, I always check the reviews. I think I can get more information about the product from the users' feedback because they provide it after using it. So, in fact, user feedback helps me decide whether to accept or reject the recommendation."

Therefore, collecting user reviews properly and systematically can help design a better user-centric recommendation system.

**Trustworthy Reviews**

We have found that most users were concerned about fake reviews that might be circulating on the internet. They were skeptical about whether the reviews were manipulated by the AI used in the recommendation system. Since a user does not know how the whole system works and there is no reliable mechanism to check if the reviews have genuinely originated from the reviewer, it is not considered trustworthy. One of the participants said,

> "It is true that I see user reviews to busy any item, but I dont know whether the review is original. I am not sure if the AI is manipulating the reviews to convince me to buy the item."

The participants concerned about the fake review said,

"The system should filter fake reviews. I do not want to make my decision based on any fake reviews".

However, the discussion focused on the trustworthiness of reviews and how various companies can encourage users to provide constructive reviews. Some participants suggested that incentives can entice the user in this regard. In addition, the participants seem to have a preconceived perception that incentive-based reviews will be trustworthy. One of the participants said,

"The companies can introduce incentives for the users to provide reviews. They can design a standard format to provide reviews. It will also help the companies to put a detailed and constructive review."

Therefore, a detailed user review can enhance the quality of the explanation. In addition, the users should also be convinced that humans do not manipulate the reviews rather, they are originally from the users.

## 5 Discussion and Implications

This work contributes to these issues from an in-depth qualitative view of 'users' requirement of the nature of explanation in the XAI system.' Our approach unfolds a set of reciprocal perspectives of recommendation systems through the lens of explainable artificial intelligence. The major implications are discussed as follows.

### 5.1 Considering End User Requirements in XAI Development

Research on XAI focuses mostly on technical perspectives such as algorithms and model interpretability for the developers and scientific community (Mohseni et al., 2021). In this paper, we have investigated the nature of user-defined explanations in recommendation systems through the lens of XAI. Our observation shows that user requirements for explanations significantly differ from person to person. Hence, the requirement for a tailor-made explanation comes to light. Meske et al. (2022) also discussed similar attributes as a quality criterion of explanations, which is confirmed in this study as one of our focus group discussion findings. Moreover, during the interview session, the participants highlighted that providing them with short training before using the explainable recommendation system is better. In this way, the users can know how the whole system works. This is one of the many features that can be implemented by involving the end user in the development process. Mohseni et al. (2021) discussed assessing user interaction in an XAI system to design and develop a human-centered system. Similarly, Ribera & Lapedriza (2019) also discussed that the end user requirement is beneficial for system development. Another finding that warrants user involvement is various shades of explanation visualization in a system. However, since there is no standard for XAI system development, the developers have no rule of thumb to follow. Though Mohseni et al. (2021) proposed a framework for XAI system development and evaluation, however, a standard guideline is yet to be defined. Therefore, designers and developers should consider the end user involvement(found in our study) to identify explanation presentation format as a design requirement for XAI development. Moreover, recent findings reveal that explanation visualizations can be personalized according to the user's demand to improve comprehensibility (Chiang & Yin, 2022; Yu et al., 2020 ). Therefore, user involvement in the development process can help the developers design a user-centric XAI system that matches the user's mental model (Mohseni et al., 2021).

## 5.2     Explanation Verifiability, and Authenticity

Users are concerned about whether the system presents them with authentic information as an explanation. This issue is more prevalent due to the widespread availability of fake product details and reviews (Wu et al., 2020). With the endeavor to promote products online, fake reviews and feedback are becoming widely prevalent in various content recommendation systems. These fake reviews can make any low-quality content highly demanding on the website, and the users can be easily fooled. To increase the reliability and trustworthiness of the system, the system must provide users with authentic reviews. Therefore, designers, developers, and company owners should be aware of this and ensure that authentic information is displayed to the users. Therefore, in line with the discussion of Mohseni et al. (2021), it is crucial to ensure the trustworthiness of the explanation. If the explanation is authentic and verifiable using proper credentials and secure methods, the trustworthiness of the explanation will increase significantly. Understandably, there is always a tradeoff between interpretability and model efficiency; however, the system should also be able to go hand in hand with the target user's demand. Miller (2019) and Gerlings (2020) also described that the XAI system introduces trust and transparency towards the system, but the issue is authentic and verifiable data as the explanation is relatively new and needs significant attention (Mohseni et al., 2021; Meske et al., 2022; Miller, 2019; Gerlings, 2020). However, the requirement for trustworthy and authentic explanations can be addressed using smart contracts and blockchain

## 5.3 Theoretical Framework for End-user Inclusion and Requirement Validation for XAI Development

In this section, we propose a theoretical framework (Figure 3) highlighting the involvement of an end user in XAI development. Our findings suggest several factors that warrant the involvement of the end user. One of the findings is to consider the end user requirements in the development of an XAI system. Since end-users are concerned about the amount of information being provided to them as an explanation. They highlighted that if any system provides a great deal of unnecessary information as an explanation, they are most likely to stop using that system. In other words, the cognitive load of explanation should be controlled and calibrated to demonstrate the usability of XAI systems (Paas et al., 2016). Agreeing with the discussion by Meske et al. (2022), to control end users' cognitive perception, we need to know how much information the system should provide them with as an explanation. Therefore, to evaluate end-user requirements their involvement is necessary. Moreover, Wang et al. (2019) also suggested that end users' involvement can be crucial to designing a trustworthy and user-friendly XAI system (Wang et el., 2019). Schoonderwoerd et al. (2021) talked about eliciting requirements to develop design patterns for XAI development (Schoonderwoerd et al., 2021). However, as elicitation techniques, they proposed designing questionnaires, establishing brainstorming or discussion with end users, etc. In our proposed framework, we plan to include end users by collecting end-user requirements using co-design, training, etc. After collecting end-user requirements, we propose to develop a framework for user requirement validation. As mentioned above, we must measure the cognitive load, trustworthiness, and usability of the requirements (Mohseni et al., 2021; Meske et al., 2022). Finally, we will be able to validate end-user requirements that can be used as a design requirement for XAI design and development. System developers can use this framework to build a more user-centric XAI system. Moreover, this framework can also be integrated with the traditional software development lifecycle to develop a comprehensive framework for XAI development.

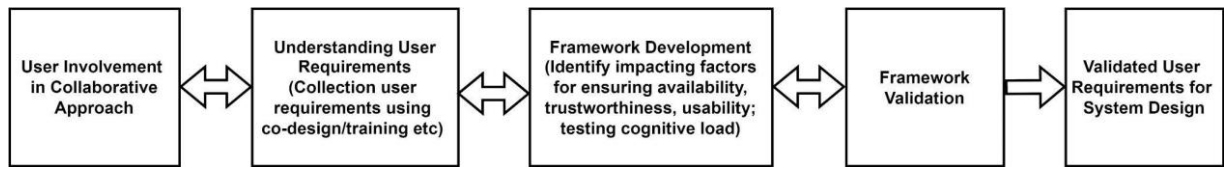

Figure 3.    User involvement in XAI Development

## 6    Limitations

This empirical study, with the participation of end-users from different backgrounds, provides the researchers with a unique opportunity to understand the human comprehension of XAI from the end-users point of view. Though our findings have brought valuable insights regarding the design and development of XAI systems for lay users.

Our empirical study is based on online content recommendation systems. We have provided three scenarios to the participants: housing recommendation, movie recommendation, and e-commerce-related product recommendation system. However, findings might differ while investigating other recommendation systems. For example, during our interview, the participants repeatedly mentioned the recommendation systems used in various social media and other news portals. Hence, empirical investigation of those recommendation systems can bring crucial implications for XAI design.

The participants of our interview session were students and young professionals aged 18 to 35 years. An empirical study of different age groups can reveal other interesting findings. Moreover, our study participants were primarily graduate students and young professionals. The results can be different for relatively low-literate people as well. In addition, the participants had less exposition to explainable artificial intelligence or such systems. The findings can be interesting if the participants are exposed to such a system for a longer period and on a daily basis. The participants were already becoming more comfortable as the interview was nearing the end. Hence, relatively long exposure to XAI systems would impact the end user's comprehension. Therefore, future research can include the exploration of XAI among various age groups, professions, and regular users of recommendation systems.

## 7      Conclusion

In this paper, we have investigated the topic of Explainable Artificial Intelligence from an end-user point of view. We have presented the users with real-life scenarios that they use and frequently interact with. This paper presents an in-depth qualitative analysis of customized and user-defined explanations that can increase the end users' interaction, usability, understandability, and trust. The findings of this work highlight the importance of lay user perception from a wider lens beyond the technical aspects of AI. Researchers and practitioners can consider the findings and discussions to identify the design requirements and solutions. Therefore, our work widens XAI's scope in terms of user-centric AI design and development. Further long-term and detailed investigation from a lay user's perspective might unfold more comprehensive XAI design requirements.